\documentclass[conference]{IEEEtran}
\IEEEoverridecommandlockouts
\usepackage{cite}
\usepackage{amsmath,amssymb,amsfonts}
\usepackage{algorithmic}
\usepackage{graphicx}
\usepackage{multirow}
\usepackage{textcomp}
\usepackage{xcolor}
\usepackage{comment}
\usepackage[hyphens]{url}
\usepackage{hyperref}
\usepackage[hyphenbreaks]{breakurl}
\def\BibTeX{{\rm B\kern-.05em{\sc i\kern-.025em b}\kern-.08em
    T\kern-.1667em\lower.7ex\hbox{E}\kern-.125emX}}
    
\ifCLASSOPTIONcompsoc
    \usepackage[caption=false, font=normalsize, labelfont=sf, textfont=sf]{subfig}
\else 
\usepackage[caption=false, font=footnotesize]{subfig}
\fi

\begin{document}

\title{An Experimental Evaluation on Deepfake Detection using Deep Face Recognition\\
}

\author{\IEEEauthorblockN{Sreeraj Ramachandran, Aakash Varma Nadimpalli, Ajita Rattani}
\IEEEauthorblockA{\textit{School of Computing, Wichita State University} \\
Wichita, USA \\
\{sxramachandran2,axnadimpalli\}@shockers.wichita.edu, ajita.rattani@wichita.edu}
}

\maketitle

\begin{abstract} 
Significant advances in deep learning have obtained hallmark accuracy rates for various computer vision applications. However, advances in deep generative models have also led to the generation of very realistic fake content, also known as deepfakes, causing a threat to privacy, democracy, and national security. Most of the current deepfake detection methods are deemed as a binary classification problem in distinguishing authentic images or videos from fake ones using two-class convolutional neural networks (CNNs). These methods are based on detecting visual artifacts, temporal or color inconsistencies produced by deep generative models. 
However, these methods require a large amount of real and fake data for model training and their performance drops significantly in cross dataset evaluation with samples generated using advanced deepfake generation techniques.  
In this paper, we thoroughly evaluate the efficacy of deep face recognition in identifying deepfakes, using different loss functions and deepfake generation techniques. Experimental investigations on challenging Celeb-DF and FaceForensics++ deepfake datasets suggest the efficacy of deep face recognition in identifying deepfakes over two-class CNNs and the ocular modality. Reported results suggest a maximum Area Under Curve (AUC) of $0.98$ and  Equal Error Rate (EER) of $7.1\%$ in detecting deepfakes using face recognition on the Celeb-DF dataset. This EER is lower by $16.6\%$ compared to the EER obtained for the two-class CNN and the ocular modality on the Celeb-DF dataset. Further on the FaceForensics++ dataset, an AUC of $0.99$ and EER of $2.04\%$ was obtained. The use of biometric facial recognition technology has the advantage of bypassing the need for a large amount of fake data for model training and obtaining better generalizability to evolving deepfake creation techniques.
\end{abstract}

\begin{IEEEkeywords}
Deepfakes, Deep Learning, Biometrics, Face Recognition
\end{IEEEkeywords}

\section{Introduction}

\begin{figure}[t!]
\centering
\includegraphics[scale=0.09]{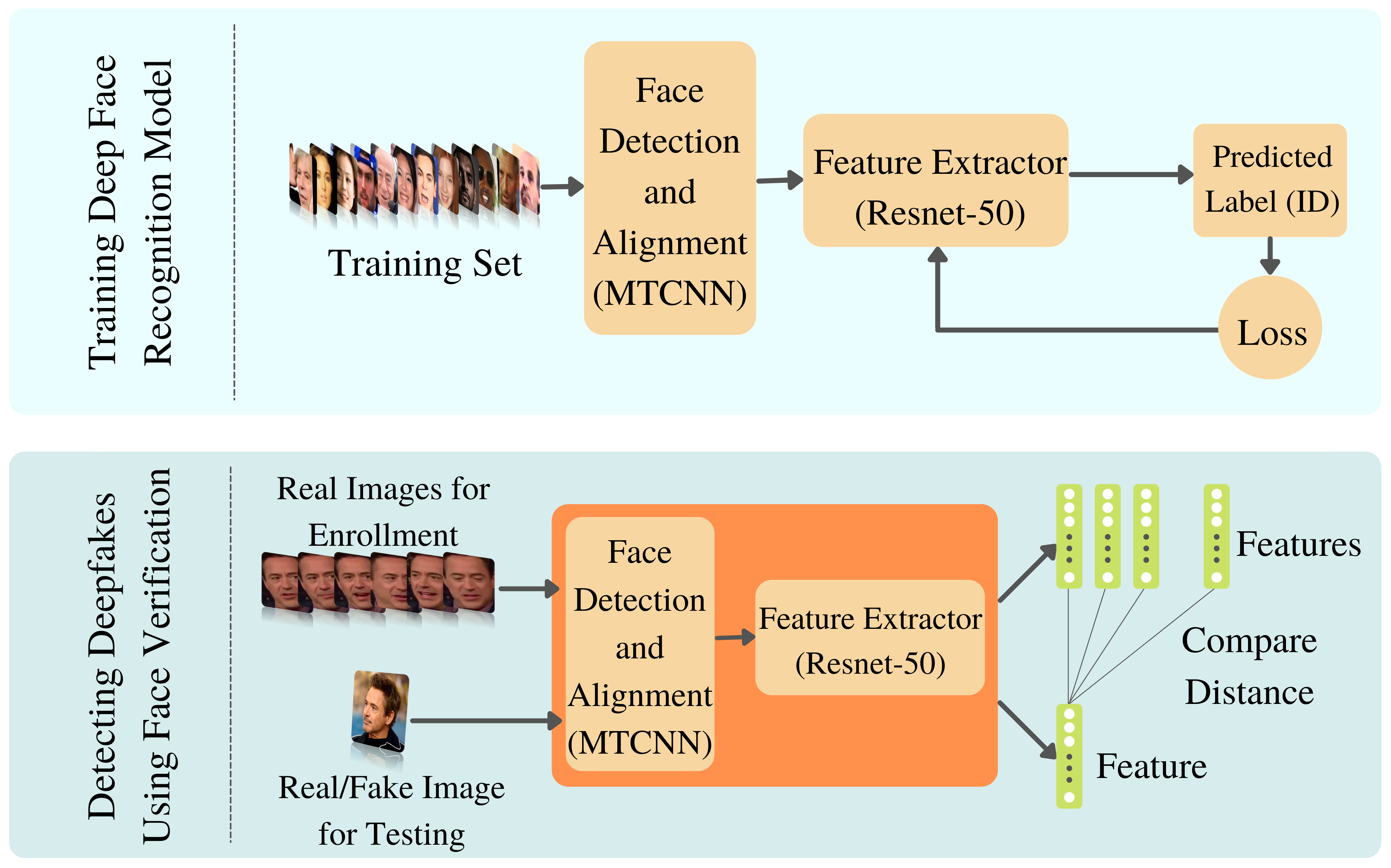}
\caption{Illustration of the Deepfake detection using deep face recognition.
The deep face recognition models pretrained on large scale facial recognition
datasets are used for detecting deep fakes. The deepfake detection is done by comparing
the authentic template of the subject with the corresponding deep fakes as in face verification.}
\label{img:pi_model}
\end{figure}
Synthesized media, particularly deepfakes, has become a major source of concern in recent years. The now known term ``deepfake" is a combination of the words ``deep learning" and ``fake" that was first introduced in late $2017$, referred to as deep adversarial models (for instance, generative adversarial networks~(GAN)~\cite{goodfellow2014generative}) that generate fake videos by \emph{swapping} a person's face with the face of another person~\cite{bbc_bitesize_2019} using methods such as FaceSwap and FaceShifter~\cite{li2020faceshifter}. Current line of research~\cite{tolosana2020deepfakes} also include \emph{expression swapping} as one of the deepfake generation techniques, such as Face2Face~\cite{thies2016face} and NeuralTextures~\cite{2019arXiv190412356T}. Since then the amount of generated media using deepfakes has increased rapidly, due to the rapid development in computer vision and more accessible and capable hardware. 
Deepfake has been a carrier of misinformation and malice since its inception, posing a political and social threat~\cite{the_guardian_2019}.

This deep learning and computer vision usage to generate high-fidelity synthetic media has thus been a major security risk and flagged as a top AI threat~\cite{Hwang2020,burt_2020}. As a result, there has been a resurgence in research into detecting facial modifications using both data-driven deep learning and biometric anti-spoofing techniques~\cite{DBLP:journals/corr/abs-1812-08685, 9211068, https://doi.org/10.1049/iet-bmt.2019.0196}.

The current CNN-based (such as ResNet, MesoInceptionNet, and XceptionNet) deepfake detection methods are mainly based on detecting visual artifacts created due to the resolution and color inconsistency between the warped face area and the surrounding context~\cite{li2020face,li2019exposing, Afchar_2018,stehouwer2019detection} during image blending operation. 
In~\cite{li2020face}, authors proposed a general method called face X-ray to detect forgery by detecting the blending boundary of a forged image using a two-class CNN model trained end-to-end using a classification loss and a loss associated with face X-ray. This model outperformed most of the existing CNN models in deepfake detection.
Studies in~\cite{9022558, 9304936} used the optical flow of facial expression as a detection cue for altered video footage detection. 


Most of the aforementioned deepfake detection methods~\cite{li2020face,li2019exposing, Afchar_2018,stehouwer2019detection,9022558, 9304936} are \emph{training-based} and therefore, obtain very high performance in intra-dataset evaluation with an AUC of about $0.99$. However, they obtain poor generalization to high-quality fakes generated due to evolving deepfake generation techniques in cross-dataset evaluation, thus, obtaining an AUC of about $0.70$.

Media articles such as~\cite{burt_2020,microsoft} suggest the \textit{use of biometric technology} in identifying deepfakes. Recently in 2020, Nguyen and Derakhshani~\cite{9211068} used ocular-based biometric matching to distinguish between real and fake images of identities using CNNs namely, lightCNN, ResNet, DenseNet and SqueezeNet trained for ocular-based user recognition. 
In~\cite{9413323}, biometric tailored loss functions~\cite{9413323} (such as, Center, ArcFace, and A-Softmax) are used for two-class CNN training for deepfake detection. 
Worth mentioning, the study in~\cite{9413323} used only \textit{face recognition tailored loss functions in CNN training} but not face recognition technology in detecting deepfakes. 

With the advances in deepfake generation techniques, \emph{visual artifacts or other inconsistencies become progressively indistinguishable} in high quality and high-resolution deepfakes, but facial features are corrupted due to the swapping and image blending operation. Therefore, face recognition may be preferable in detecting fake images by matching corrupted feature vectors of the swapped face with the original templates of the target identity. 
A minimal analysis was carried out in ~\cite{korshunov2018deepfakes} on a first generation dataset using 129 measures of face image quality which include signal to noise ratio, specularity, blurriness, etc., along with Support Vector Machine for deepfake detection. Studies in~\cite{agarwal_protecting_2019, 9360904, DBLP:journals/corr/abs-2012-03930, DBLP:journals/corr/abs-2012-02512} have used behavioral biometrics i.e., facial expression, head, and body movement in deepfake detection. 
These studies performed limited evaluation on identity swapping based deep fake generation techniques~\cite{agarwal_protecting_2019, 9360904, DBLP:journals/corr/abs-2012-03930, DBLP:journals/corr/abs-2012-02512}. 

This paper aims to evaluate the efficacy of \emph{deep face recognition\footnote{The term face recognition and face verification are used interchangeably in this study.}, trained using different loss functions, in identifying deepfakes generated using various methods}. 
This study will provide us \emph{insight} into which deepfake generation techniques could be effectively detected by face recognition technology and the optimum loss functions to be used. 
This work has the advantage of bypassing the need for model training on fake images. However, the identity labels are required for biometric comparison (template and query image comparison) in deepfake detection. Figure~\ref{img:pi_model} illustrates deepfake detection using deep face recognition.

In summary, the main contributions of this paper are as follows: 
\begin{itemize}
  \item To evaluate the \textbf{efficiency of face recognition} in identifying deepfakes generated using various methods. To this aim, deep face recognition model based on ResNet-50 pretrained on large-scale facial recognition datasets namely, MS1M-ArcFace~\cite{deng2019arcface} and WebFace~\cite{zhu2021webface260m} using \textbf{six different loss functions} (Softmax, ArcFace, Combined Loss, SphereFace, CosFace, and Triplet loss) are used for deepfake detection.    
  
  \item Use of \textbf{explainable AI} based t-distributed stochastic neighbor embedding (t-SNE)~\cite{JMLR:v9:vandermaaten08a} in visualizing the effectiveness of facial recognition technology in detecting deepfakes.
  
  \item Thorough experimental investigation on commonly adopted CelebDF~\cite{Celeb_DF_cvpr20} and FaceForensics++~\cite{rossler2019faceforensics++} datasets across \textbf{five} different deepfake detection methods namely, FaceSwap, Face2face, FaceShifter, NeuralTextures and Deepfakes\footnote{The method of using autoencoders in deepfake generation is simply termed as DeepFakes in existing literature~\cite{tolosana2020deepfakes, Mirsky_2021}.}. 
  \end{itemize}

This paper is organized as follows: Section II discusses the methods for deepfake generation. Section III presents the dataset and the experimental protocol. Results and discussion are presented in section IV. Conclusions and future work are discussed in section V.

\section{Methods for Deepfake Generation}
In this section, we will discuss various methods used for deepfake creation~\cite{tolosana2020deepfakes, Mirsky_2021}. Depending on the level of manipulation, these deepfake creation methods can be broadly categorized as \textit{Identity Swap} and \textit{Expression Swap}~\cite{tolosana2020deepfakes}. 
\begin{enumerate}
  
    \item \textbf{Identity Swap}: Consists of replacing the face of a person in a video with the face of another person. FaceSwap\footnote{https://github.com/deepfakes/faceswap}, FaceShifter~\cite{li2020faceshifter}, and Deepfakes~\cite{DBLP:journals/corr/abs-2005-05535} are an example of identity swapping methods for deepfake creation explained as follows:   
\begin{enumerate}

\item \textbf{FaceSwap}: An identity swapping method that transfers the face region from a source video to a target video using a graphics-based approach based on detected facial landmarks. To swap the face of the source person to the target person, it uses face alignment, Gauss-Newton optimization, and image blending.
 
\item \textbf{FaceShifter}: FaceShifter~\cite{li2020faceshifter} is an identity swapping method that uses a new generator with Adaptive Attentional Denormalization (AAD) layers utilized to adaptively integrate the identity and the attributes for face synthesis, and an attributes encoder used to extract multi-level target face attributes. 

\item \textbf{Deepfakes}: In this method, two autoencoders that share the encoder are trained to reconstruct the source and target faces. To generate the fake image, the source face's trained encoder and decoder are applied to the target face, and the autoencoder's output is then blended using Poisson image editing.
\end{enumerate}
   
    \item \textbf{Expression Swapping}: Consists of facial reenactment by modifying the facial expression of the person. 
    
\begin{enumerate}
\item \textbf{Face2Face}: In Face2Face~\cite{thies2016face}, a temporary face identity is established using the first few frames, and then the expression is tracked over the remaining frames. Then, by transferring the source expression parameters of each frame to the target video, fake videos were created.

\item \textbf{NeuralTextures}: NeuralTextures~\cite{2019arXiv190412356T} employs a rendering network with a patch-based GAN-loss, to learn a neural texture of the target individual from the source video. Only the mouth-related facial expression is altered, making it extremely difficult to detect.

\end{enumerate}
\end{enumerate}

\section{Dataset and Experiment Protocol}
In this section, we discuss the datasets used and the experimental protocol followed for deepfake detection using face recognition technology.

\subsection{Datasets}
In order to evaluate the efficiency of face recognition technology in identifying high-quality deepfakes without visual artifacts, generated using various methods. 
We tested our model on the high quality deepfake datasets namely Celeb-DF~\cite{Celeb_DF_cvpr20} and FaceForensics++~\cite{rossler2019faceforensics++} with most realistic fake videos. These commonly adopted high quality deepfake datasets also have \emph{subject identity labels which facilitates} face recognition.

These datasets are described as follows:

\begin{itemize}
   
 \item \textbf{Celeb-DF}: The Celeb-DF~\cite{Celeb_DF_cvpr20} deepfake forensic dataset include $590$ genuine videos from $59$ celebrities as well as $5639$ deepfake videos. Celeb-DF, in contrast to other datasets, has essentially no splicing borders, color mismatch, and inconsistencies in face orientation, among other evident deepfake visual artifacts.

 \item \textbf{FaceForensics++}: FaceForensics++~\cite{rossler2019faceforensics++} is an automated benchmark for facial manipulation detection. It consists of several manipulated videos created using two different categories: Identity Swapping (FaceSwap, FaceSwap-Kowalski, FaceShifter, Deepfakes) and Expression swapping (Face2Face, and NeuralTextures). 
 We used the FaceForensics++ dataset's $C23$ compressed test set, which has a curated list of $70$ videos for each of these deepfake creation methods.

\end{itemize}

\subsection{Experimental Protocol}
We used the popular ResNet~\cite{he2015deep} architecture as it is widely used for face recognition. ResNet is a short form of residual network based on the  idea  of  ``identity  shortcut  connection''  where input  features  may  skip  certain  layers~\cite{he2015deep}.  In this study, we used ResNet-$50$ which has $23.5$M parameters. ResNet-50~\cite{he2015deep} is trained from scratch on MS1M-Arcface~\cite{guo2016msceleb1m, deng2019arcface} and WebFace12M dataset~\cite{zhu2021webface260m}. The MS1M-Arcface dataset~\cite{deng2019arcface}, is a cleaned version of the MS1M dataset~\cite{guo2016msceleb1m}, containing around $5.8$ million images from $85,742$ subjects. The WebFace12M  dataset~\cite{zhu2021webface260m}, is a cleaned version and subset of the complete WebFace260M dataset~\cite{zhu2021webface260m}, which contains around $12M$ face images from $600K$ identities.  The face images were detected and aligned using MTCNN~\cite{Zhang_2016}. MTCNN utilizes a cascaded CNNs based framework for joint face detection and alignment. The images are then resized to $112\times112$ for both training and evaluation.
The model was trained using \emph{six different loss functions} i.e.,  ArcFace~\cite{deng2019arcface}, CosFace~\cite{wang2018cosface}, SphereFace~\cite{wang2018cosface}, Combined Margin~\cite{deng2019arcface}, Triplet loss~\cite{deng2019arcface} and SoftMax. ArcFace, SphereFace and CosFace loss functions learn intra-class similarity and inter-class diversity for performance enhancement.

For the ResNet-$50$ model, the batch-normalization layer followed by the last fully connected layer of size $512$ and the final output layer equal to the number of classes (subjects) were used. The angular margin penalty hyper-parameters $m_1$, $m_2$ and $m_3$  was set to $1$, $0.3$ and $0.2$ for the combined margin. The networks were trained using a Stochastic Gradient Descent~(SGD) optimizer with a batch size of $64$ for $25$ epochs. The learning rate was set equal to $0.1$ at the onset and was divided at $100K$, $160K$ iterations, and finish at $180K$ iterations following the original ArcFace~\cite{deng2019arcface} implementation. We also set the momentum to $0.9$ and weight decay to $5e-4$. All the experiments are conducted on an Intel Xeon processor and two Nvidia Quadro RTX $6000$ GPUs. 

For the subject-disjoint evaluation of the pretrained deep learning face recognition models for deepfake detection on Celeb-DF~\cite{Celeb_DF_cvpr20} and FaceForensics++~\cite{rossler2019faceforensics++}, the gallery set consists of $20$ real face images (frames extracted from videos) per subject. \emph{Subject-disjoint protocol means that the subject identities used in the training face recognition model do not overlap with identities used for deepfake evaluation}. Recall that MS1M-ArcFace and WebFace12M datasets were used for training ResNet-50 based face recognition model.
The probe set consists of $1000$ real and fake face images randomly selected per subject. Depending on the dataset’s multiple videos per subject (CelebDF, DFD (subset of FF++)) or single video per subject (FF++) are used to create pairs of frames. From the gallery and probe set of all the subjects, deep features of size $512$ are extracted using the pretrained ResNet-50 model. 
The genuine match score is calculated between frames extracted from real videos and gallery face images of the corresponding subject, and the imposter score is calculated between frames extracted from fake videos and gallery face images of the target identity using \emph{cosine similarity metrics}. 
We used Equal Error Rate (EER) and Area Under the Curve (AUC) as the performance metrics for the deepfake detection using face recognition technology. 

\section{Results and Discussions}

\begin{figure*}[!bt] 
    \centering
  \subfloat[FaceSwap\label{1a}]{%
       \includegraphics[width=0.3\linewidth]{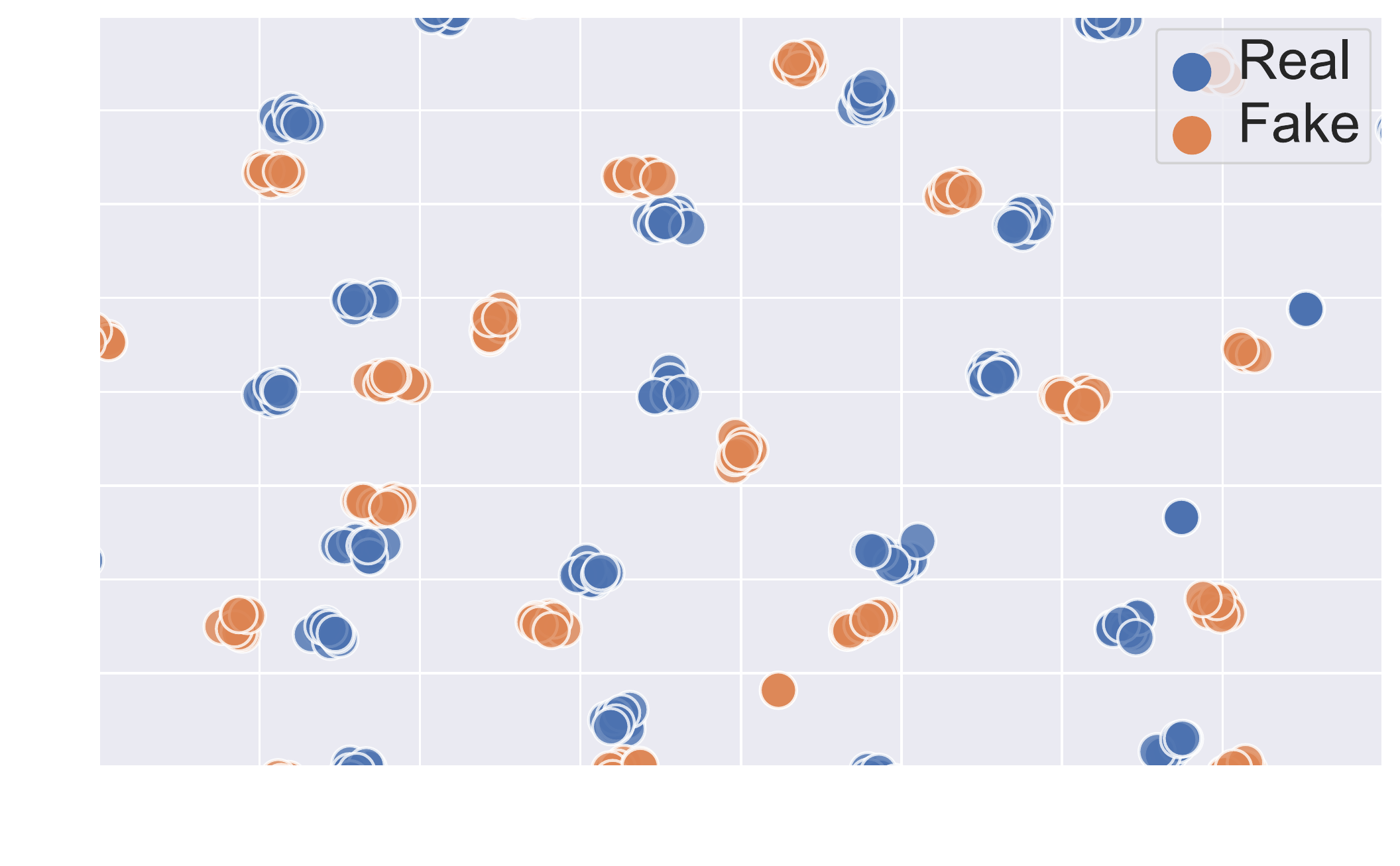}} 
        \hfill
        \subfloat[Deepfakes\label{1f}]{%
        \includegraphics[width=0.3\linewidth]{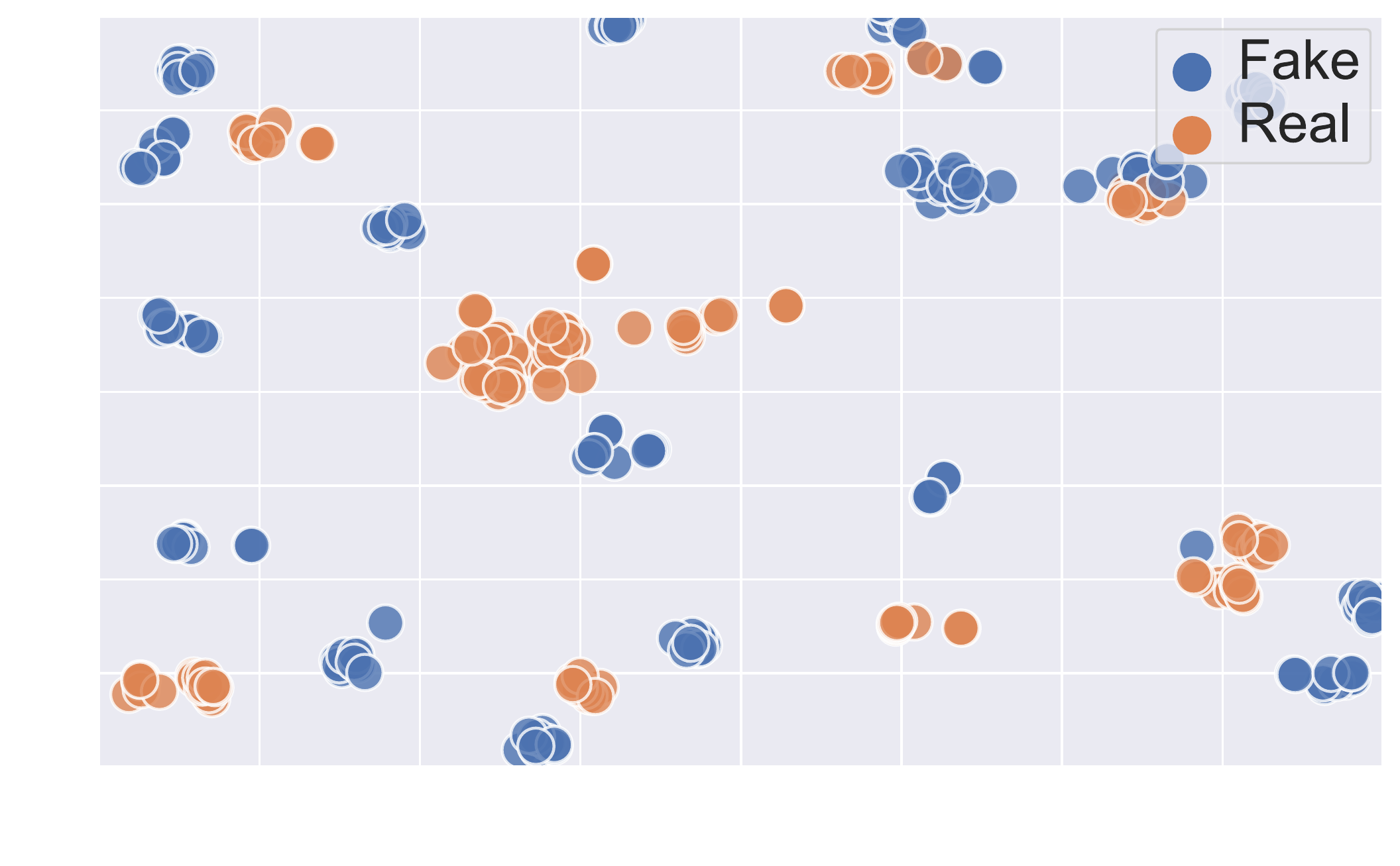}}
        \hfill
        \subfloat[NeuralTextures\label{1e}]{%
        \includegraphics[width=0.3\linewidth]{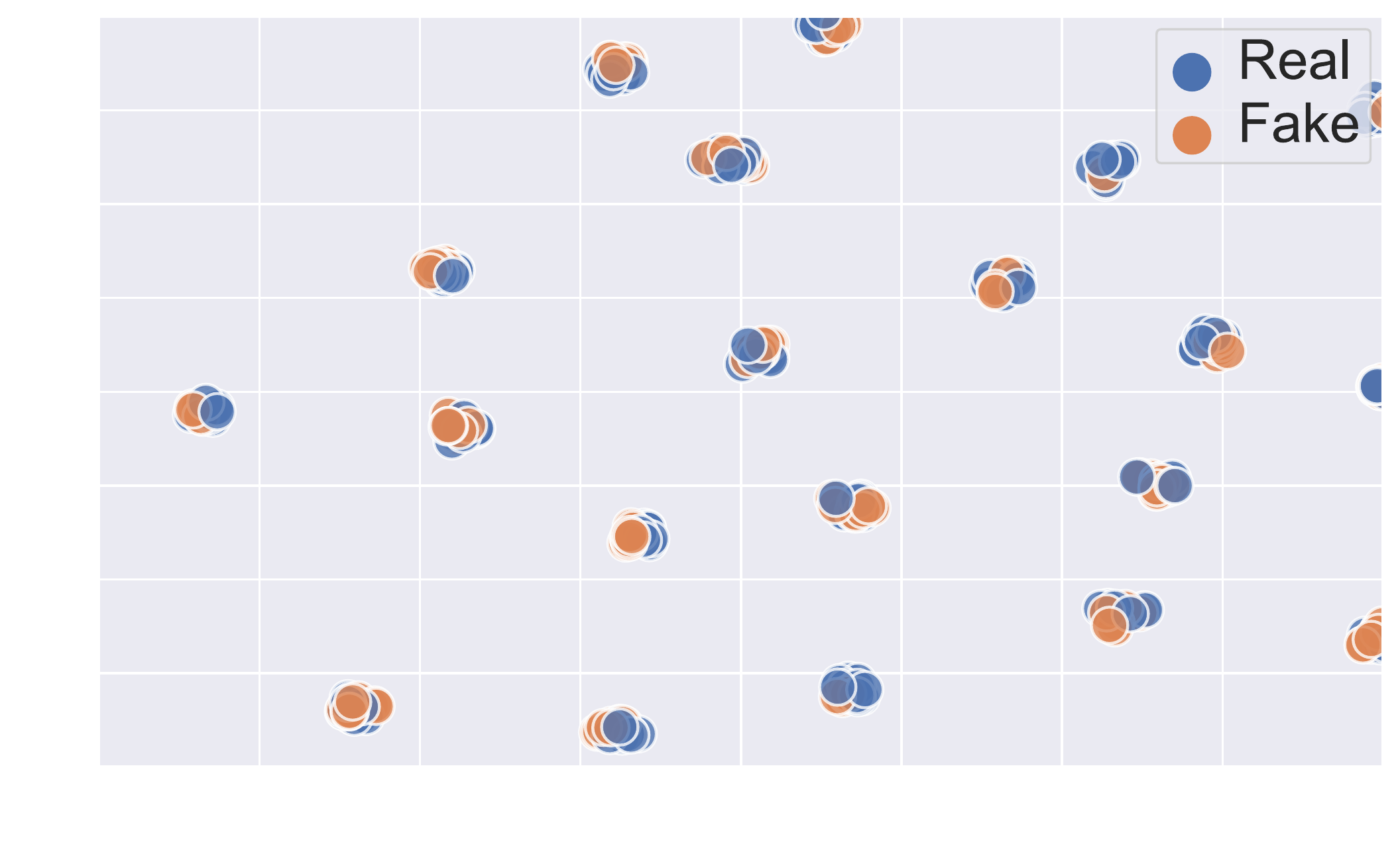}}

  \caption{t-SNE visualization of the deep feature embedding of genuine and fake images for FaceSwap, DeepFakes and NeuralTextures techniques for randomly selected identities on MS1M-AF~\cite{deng2019arcface} dataset. Similar observation was noted for other identity swapping and expression methods and on WebFace12M~\cite{zhu2021webface260m} dataset.} 
  \label{fig2} 
\end{figure*}

\begin{figure*}[!bt] %
    \centering
  \subfloat[FaceSwap\label{2a}]{%
       \includegraphics[width=0.30\linewidth]{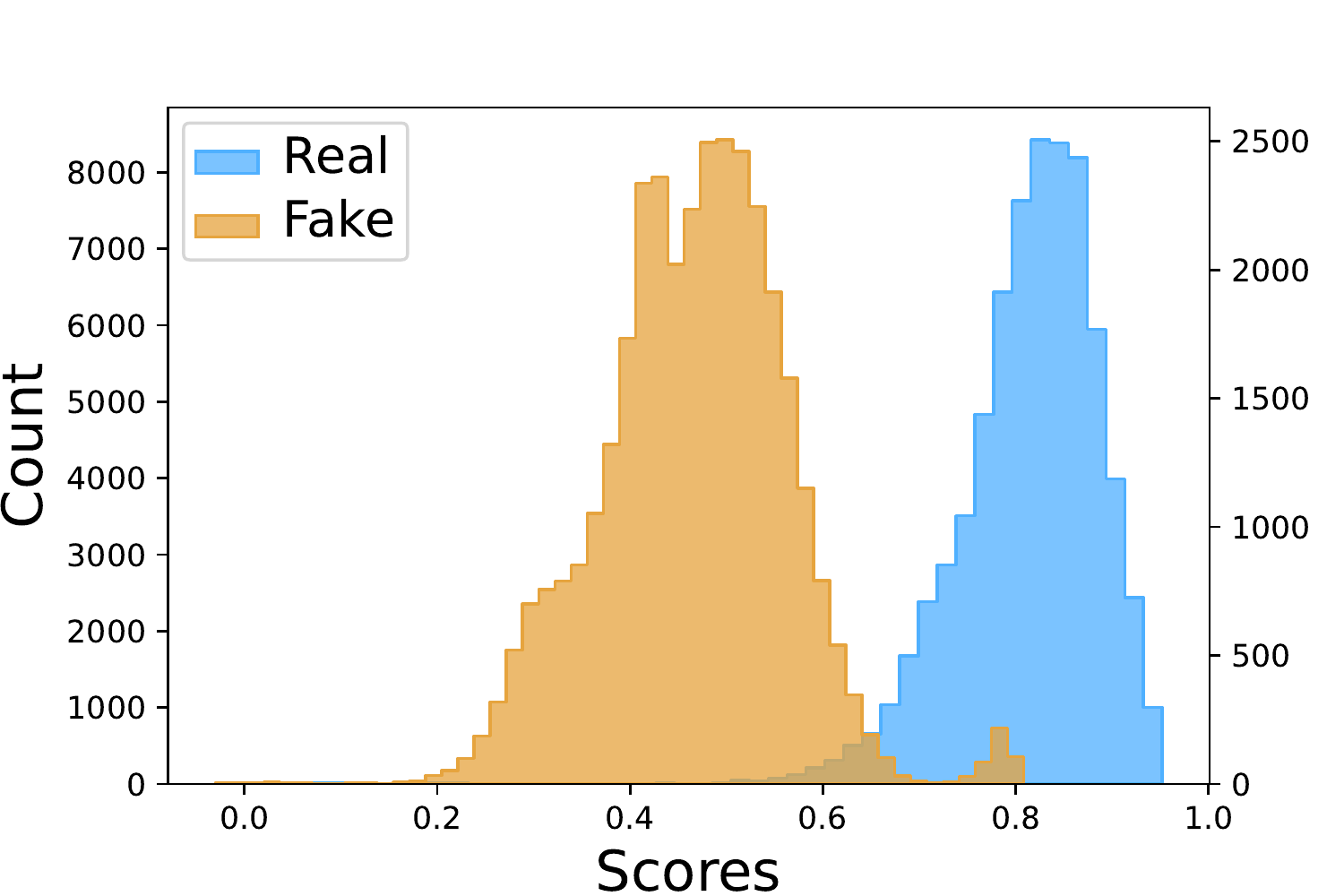}}
    \hfill
 \subfloat[Deepfakes\label{2f}]{%
        \includegraphics[width=0.30\linewidth]{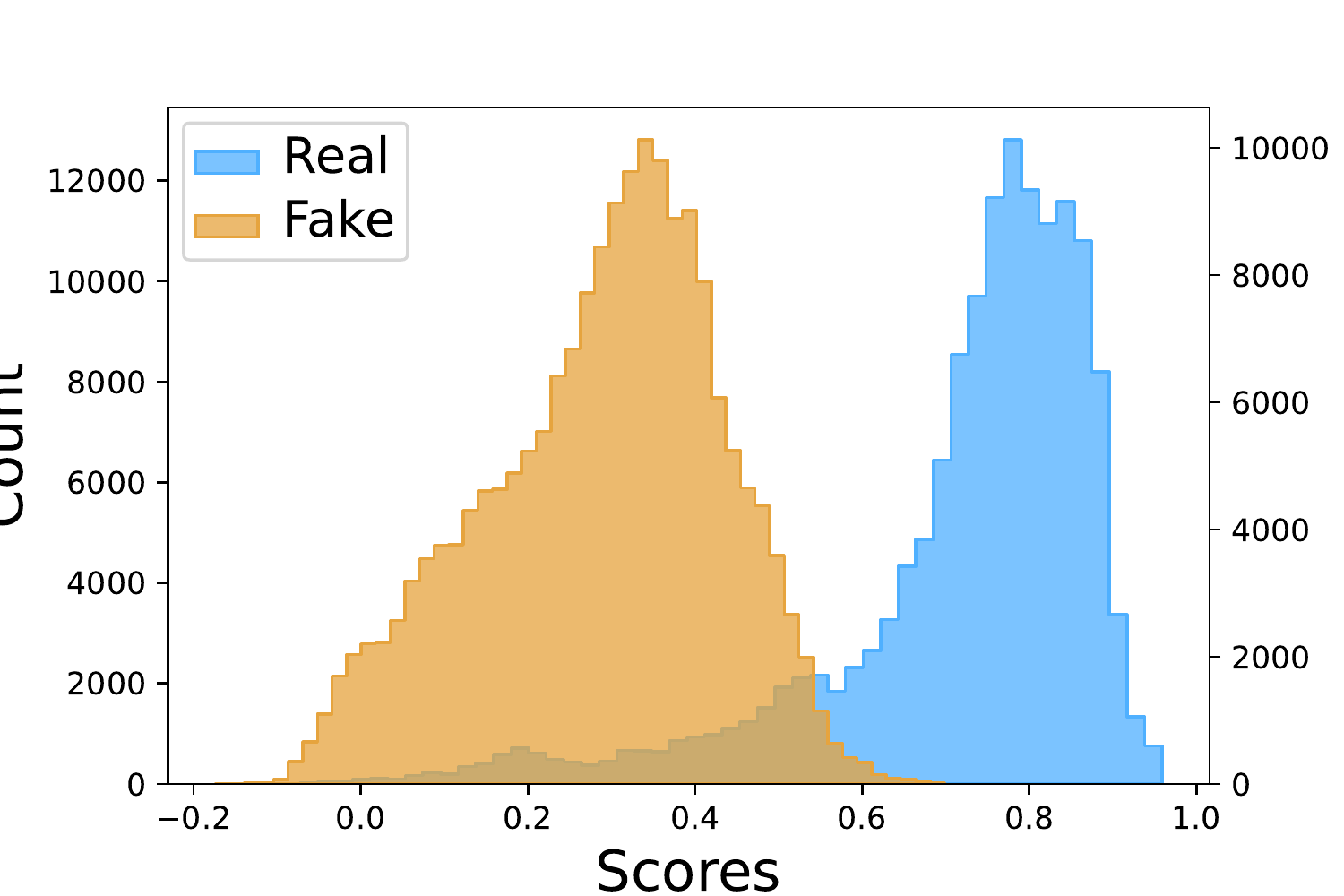}}
        \hfill
  \subfloat[NeuralTextures\label{2e}]{%
        \includegraphics[width=0.30\linewidth]{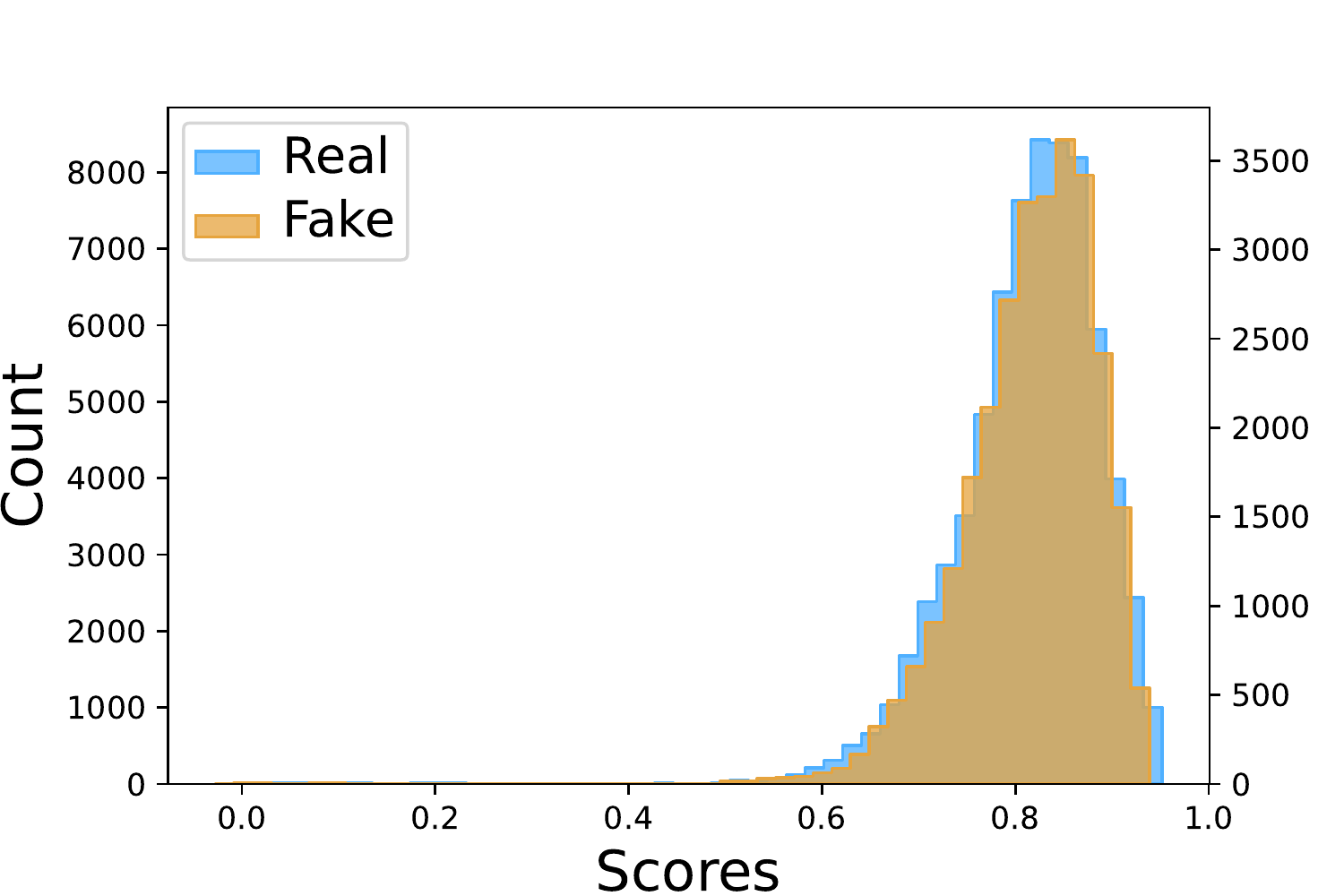}}
  \caption{The genuine and deepfake score distribution obtained on comparing the templates with the genuine query and deepfakes, respectively, for FaceSwap, Deepfakes and NeuralTextures techniques on MS1M-AF~\cite{deng2019arcface} dataset. Similar observation was noted for other identity swapping and expression methods and on WebFace12M~\cite{zhu2021webface260m} dataset.} 
  \label{fig3} 
\end{figure*}

\begin{table*}[!h]
	\begin{center}
		\caption{The AUC and EER scores obtained on using ResNet-50 face recognition model in detecting deepfakes generated using various techniques on high quality Celeb-DF~\cite{Celeb_DF_cvpr20} and FaceForensics++~\cite{rossler2019faceforensics++} datasets. These results are included for both MSIM-AF~\cite{deng2019arcface} and WebFace12M~\cite{zhu2021webface260m} datasets used for training the ResNet-50 based face recognition model. Comparison has been done with the existing methods based on ocular region~\cite{9211068} and the best performing face-Xray based CNN model~\cite{li2020face} for deepfake detection.}
		\scalebox{0.95}{
			\begin{tabular}{|cc|c|c|c|c|c|c|c|c|c|c|c|c|c|c|}
				\hline
				&                         & \multicolumn{2}{c|}{\textbf{Celeb-DF}}                                                                                                                                                                                                                                         & \multicolumn{12}{c|}{\textbf{FaceForensics++}} \\ \cline{3-16}
				&   & \multicolumn{10}{c|}{\textbf{Identity Swapping Methods}} & \multicolumn{4}{c|}{\textbf{Expression Swapping Methods}} \\ \cline{1-16}
				
				\multicolumn{1}{|c|}{}                      & \multirow{2}{*}{Loss}   & \multicolumn{2}{c|}{Deepfakes}              & \multicolumn{2}{c|}{Deepfakes}              & \multicolumn{2}{c|}{FaceShifter}           & \multicolumn{2}{c|}{FaceSwap-K}            & \multicolumn{2}{c|}{FaceSwap}         & \multicolumn{2}{c|}{Face2Face}             & \multicolumn{2}{c|}{NeuralTextures}                          \\ \cline{3-16} 
				\multicolumn{1}{|c|}{}                                                               &                     & \textbf{AUC}         & \textbf{EER}        & \textbf{AUC}         & \textbf{EER}        & \textbf{AUC}         & \textbf{EER}        & \textbf{AUC}         & \textbf{EER}        & \textbf{AUC}           & \textbf{EER}        & \textbf{AUC}         & \textbf{EER}         & \textbf{AUC}         & \textbf{EER}         \\ \hline
				\multicolumn{1}{|c|}{\multirow{6}{*}{\rotatebox[origin=c]{90}{\textbf{MS1M-AF}}}}    & \textbf{Softmax}    & $0.938$              & $13.22$             & $0.928$              & $13.56$             & $0.916$              & $16.29$             & $0.985$              & $3.97$              & $0.973$                & $7.57$              & $0.774$              & $30.34$              & $\boldsymbol{0.517}$ & $\boldsymbol{48.37}$ \\
				\multicolumn{1}{|c|}{}                                                               & \textbf{Arcface}    & $0.964$              & $9.07$              & $0.943$              & $10.70$             & $0.962$              & $9.50$              & $0.989$              & $2.40$              & $0.985$                & $4.23$              & $0.746$              & $33.03$              & $0.501$              & $49.53$              \\
				\multicolumn{1}{|c|}{}                                                               & \textbf{Combined}   & $0.976$              & $7.47$              & $0.954$              & $9.24$              & $0.975$              & $6.53$              & $0.99$               & $2.09$              & $0.989$                & $2.89$              & $0.737$              & $33.79$              & $0.49$               & $50.74$              \\
				\multicolumn{1}{|c|}{}                                                               & \textbf{SphereFace} & $0.954$              & $10.63$             & $0.948$              & $10.67$             & $0.957$              & $10.99$             & $0.984$              & $3.50$              & $0.982$                & $4.61$              & $\boldsymbol{0.801}$ & $\boldsymbol{27.61}$ & $0.516$              & $48.72$              \\
				\multicolumn{1}{|c|}{}                                                               & \textbf{CosFace}    & $\boldsymbol{0.98}$  & $\boldsymbol{7.10}$ & $\boldsymbol{0.959}$ & $\boldsymbol{8.38}$ & $\boldsymbol{0.979}$ & $\boldsymbol{5.15}$ & $\boldsymbol{0.99}$  & $\boldsymbol{2.04}$ & $\boldsymbol{0.99}$    & $\boldsymbol{2.68}$ & $0.721$              & $35.55$              & $0.485$              & $51.10$              \\
				\multicolumn{1}{|c|}{}                                                               & \textbf{Triplet}    & $0.793$              & $28.04$             & $0.879$              & $20.75$             & $0.927$              & $13.66$             & $0.957$              & $8.10$              & $0.913$                & $14.49$             & $0.641$              & $39.99$              & $0.472$              & $52.04$              \\ \hline
				\multicolumn{1}{|c|}{\multirow{6}{*}{\rotatebox[origin=c]{90}{\textbf{WebFace12M}}}} & \textbf{Softmax}    & $0.934$              & $12.47$             & $0.912$              & $15.72$             & $0.908$              & $17.48$             & $0.972$              & $4.97$              & $0.954$                & $8.54$              & $\boldsymbol{0.787}$ & $\boldsymbol{31.35}$ & $\boldsymbol{0.509}$ & $\boldsymbol{49.23}$ \\
				\multicolumn{1}{|c|}{}                                                               & \textbf{Arcface}    & $0.954$              & $10.76$             & $0.932$              & $13.43$             & $0.952$              & $10.42$             & $0.976$              & $3.84$              & $0.979$                & $6.43$              & $0.754$              & $34.21$              & $0.500$              & $50.28$              \\
				\multicolumn{1}{|c|}{}                                                               & \textbf{Combined}   & $\boldsymbol{0.975}$ & $\boldsymbol{8.21}$ & $0.943$              & $10.13$             & $0.972$              & $7.92$              & $\boldsymbol{0.982}$ & $\boldsymbol{2.96}$ & $\boldsymbol{0.982}$   & $\boldsymbol{4.97}$ & $0.741$              & $34.78$              & $0.489$              & $51.21$              \\
				\multicolumn{1}{|c|}{}                                                               & \textbf{SphereFace} & $0.944$              & $11.57$             & $0.938$              & $11.98$             & $0.914$              & $12.49$             & $0.962$              & $4.59$              & $0.975$                & $6.78$              & $0.778$              & $31.78$              & $0.491$              & $50.06$              \\
				\multicolumn{1}{|c|}{}                                                               & \textbf{CosFace}    & $0.973$              & $9.23$              & $\boldsymbol{0.947}$ & $\boldsymbol{9.48}$ & $\boldsymbol{0.974}$ & $\boldsymbol{6.83}$ & $0.971$              & $3.07$              & $0.980$                & $4.98$              & $0.719$              & $37.89$              & $0.483$              & $52.45$              \\
				\multicolumn{1}{|c|}{}                                                               & \textbf{Triplet}    & $0.812$              & $30.42$             & $0.884$              & $22.17$             & $0.903$              & $15.25$             & $0.942$              & $9.23$              & $0.899$                & $16.34$             & $0.624$              & $41.97$              & $0.467$              & $54.23$              \\ \hline
				\multicolumn{2}{|c|}{\parbox{2cm}{\textbf{Nguyen and Derakhshani~\cite{9211068} (Ocular Recognition, Softmax)}}} & $\boldsymbol{0.879}$                  & $\boldsymbol{20.70}$                 & $-$                  & $-$                  & $-$                  & $-$                 & $-$                  & $-$                 & $-$                  & $-$                  & $-$                  & $-$                 & $-$                & $-$                \\ 
				\hline
				\multicolumn{2}{|c|}{\parbox{2cm}{\textbf{Li et al.~\cite{li2020face} (Face-Xray, Softmax)}}} & $\boldsymbol{0.806}$                  & $\boldsymbol{26.70}$                 & $-$                  & $-$                  & $-$                  & $-$                 & $-$                  & $-$                 & $-$                  & $-$                  & $-$                  & $-$                 & $-$                & $-$                \\ \hline
			    
																								
			\end{tabular}
		}
		\label{tab1}
	\end{center}
\end{table*}

Table~\ref{tab1} shows the deepfake detection performance in terms of Equal Error Rate (EER) and AUC evaluated on Celeb-DF and FaceForensics++ datasets, using face recognition ResNet-50 model pretrained on MS1M-ArcFace and WebFace12M dataset. The top performance results are highlighted in bold for both the training datasets and across the different deepfake generation techniques.

As can be seen from the table, \textbf{Identity swapping methods} namely, FaceSwap, FaceShifter, and Faceswap-Kowalski obtain AUC close to $1$ and EER of $3.29\%$ on an average, demonstrating the efficiency of face recognition in detecting fakes created using these methods. The deepfake methods based on autoencoders also obtained a high AUC score of about $0.97$. \emph{This demonstrates that even high-quality deepfakes without apparent visual artifacts, such as those in Celeb-DF datasets, have their facial features corrupted using blending operation used in face-swapping techniques}. These corrupted facial features are efficiently detected using a face recognition algorithm by matching original templates to the deepfake images of the target identity. The obtained results using face recognition technology significantly outperformed the ocular recognition for deepfake detection~\cite{9211068} by $13.6\%$ using subject-disjoint evaluation on CelebDF. The obtained results are also better than the popular two-class CNN-based Face-Xray model~\cite{li2020face} by $19.6\%$ on the Celeb-DF dataset (see Table~\ref{tab1}). The Face-Xray model was chosen for comparison because it outperformed other CNN models in~\cite{li2020face}. \textbf{N}ote that the existing studies~\cite{agarwal_protecting_2019, 9360904, DBLP:journals/corr/abs-2012-03930, DBLP:journals/corr/abs-2012-02512} on using behavioral biometrics (such as facial expression and head-pose movement) for deepfake detection, performed a very limited evaluation on identity swapping based deep fake generation techniques and reported only AUC score. Therefore, we did not use these studies for cross-comparison. 

\textbf{Expression swapping methods} namely, Face2Face and NeuralTextures obtained the least performance with EER of $27.61\%$ and $48.37\%$, respectively, for the best case. The obtained EER is about ~$32.5\%$ higher than those obtained for identity swapping techniques. The NeuralTextures approach further obtained lower performance over Face2Face, because it primarily altered the face expression that corresponded to the mouth region resulting in less deformed facial discriminators. Face2Face uses a re-targeting and warping procedure to swap expressions, leaving many of the original facial features intact, resulting in the facial recognition model failing in detecting deepfakes. \emph{Thus, facial recognition technology is not effective in detecting deepfake generated using expression swapping methods. This is primarily due to the fact that only expression is changed keeping the identity features intact.} Experiments on the FF++ dataset suggest that even within the same environment and conditions (i.e. by using single video per subject), expression-swapping methods severely underperform.
The experimental results also suggest that the dataset used for training face recognition model (i.e., MS1M-ArcFace and WebFace12M) has no impact on deepfake detection accuracy evaluated on Celeb-DF and FaceForensics++ deepfake datasets. 

Among different \textbf{loss functions} that were used to train the face recognition model, \textit{Combined margin and CosFace loss} performed the best by about $3.43\%$, over other loss functions. Triplet loss performed the least on average. When trained on MS1M-Arcface dataset, CosFace obtained the best performance on detecting FaceSwap ($0.99$ AUC, $2.68\%$ EER), FaceShifter ($0.98$ AUC, $2.04\%$ EER), FaceSwap-Kowalski ($0.99$ AUC, $2.04\%$ EER), and Deepfakes ($0.959$ AUC, $8.38\%$ EER) in Faceforensic++ and Celeb-DF ($0.98$ AUC, $7.1\%$ EER) dataset. Using SphereFace, the best performance ($0.801$ AUC, $27.61\%$) was obtained in detecting Face2Face and NeuralTextures were best detected  ($0.517$ AUC, $48.37\%$ EER) using Softmax loss. 

Similarly, when trained on WebFace12M dataset, Combined Margin obtained the best performance on detecting FaceSwap ($0.982$ AUC, $4.97\%$ EER), FaceSwap-Kowalski ($0.982$ AUC, $2.96\%$ EER), Deepfakes ($0.947$ AUC, $9.48\%$ EER) for faceforensics++ and Celeb-DF ($0.975$ AUC, $8.21\%$ EER) dataset. Face2Face ($0.787$ AUC, $31.35\%$ EER) and NeuralTextures ($0.509$ AUC, $49.23\%$ EER) were best detected using  Softmax and FaceShifter ($0.974$ AUC, $6.83\%$ EER) using CosFace loss function. Loss functions based on \emph{Cosine Margin Penalty}, such as ArcFace, CosFace, and Combined Margin, fared better on an average by $9.3\%$ than the others in deepfake detection, whereas triplet loss obtained the least performance.

Figure~\ref{fig2} shows the t-SNE~\cite{JMLR:v9:vandermaaten08a} visualization of the deep feature embeddings (number of components = $2$ for real and fake images) from FaceSwap, NeuralTextures and Deepfakes based fake creation techniques. There is a clear separation between facial features from genuine and deepfakes for the FaceSwap approach. A similar observation was obtained for other identity swapping methods as well. However, the real and fake features overlap when using expression swapping approaches, as can be seen for the NeuralTexture method.

Another noteworthy finding from these visualizations is that for Deepfakes, which are based on auto-encoders, deep features are considerably more spread out than the other approaches indicating the lack of consistency across different frames of the same subject. Figure~\ref{fig3} shows the histogram of the genuine and deepfake scores distribution obtained on comparing the templates with real and fake images using different deepfake generation techniques. Similar to t-SNE based visualization, expression swapping methods obtained higher overlap in genuine and deepfake distribution over identity swapping methods.

In \textbf{summary}, experimental results demonstrate the effectiveness of face recognition technology in identifying different identity swapping-based deepfake generation methods. Combined margin and CosFace loss functions obtained the best deepfake detection rate as they can attain better intra-class compactness as well as can maximize inter-class separability.
The obtained results using face recognition technology significantly outperformed the existing biometric studies using facial region~\cite{9211068} and the popular Face-Xray model~\cite{li2020face} for deepfake detection (see Table~\ref{tab1}). The facial recognition technology is not effective in detecting deepfake generation using expression swapping methods that only change expression keeping the identity features intact.

\section{Conclusion and Future Work}
 As most of the existing deepfake detection algorithms rely on visible structural artifacts or color inconsistencies, they do not perform well on high-quality datasets comprising next-generation deepfakes, such as those available in Celeb-DF and FaceForensics++. In this paper, we evaluated the effectiveness of deep face recognition in detecting high-quality deepfake images or videos from the real ones of the same identity, using the notion of detecting corrupted facial features rather than image anomalies. Experimental results demonstrated the efficiency of face recognition technology in identifying deepfake based identity swapping methods, surpassing those obtained by two-class CNNs on the same datasets. Combined margin and CosFace loss functions obtained the best deepfake detection rate. However, the face recognition technology could not be used for detecting expression swaps in deepfakes. One of the limitations of using biometric technology for detecting deepfakes is the requirement of the subject's identity for biometric facial feature matching. As a part of future work, the bias of face recognition technology in identifying deepfakes across demographic variations will be evaluated. Through investigation will be done to understand the effectiveness and failure mode of face recognition technology across evolving deepfake generation techniques. The fusion of behavioral biometrics with facial features will be explored for enhanced deepfake detection performance.

\section{Acknowledgement}
This work is supported in part from a grant no. \#210716 from University Research/Creative Projects at Wichita State University. The research infrastructure is supported in part from a grant No. 13106715 from the Defense University Research Instrumentation Program (DURIP) from Air Force Office of Scientific Research.

\bibliographystyle{./IEEEtran}
\bibliography{./IEEEabrv,./mybibfile}

\end{document}